\documentclass[10pt,twocolumn,letterpaper]{article}

\usepackage{iccv}
\usepackage{times}
\usepackage{epsfig}
\usepackage{graphicx}
\usepackage{amsmath}
\usepackage{amssymb}
\usepackage{booktabs}
\usepackage{bm}
\usepackage[table]{xcolor}
\usepackage{multirow}
\usepackage{multicol}
\usepackage{makecell}
\usepackage{listings}
\usepackage{appendix}
\usepackage{authblk}
\usepackage{subfig}

\usepackage[breaklinks=true,bookmarks=false]{hyperref}

\iccvfinalcopy % *** Uncomment this line for the final submission

\def\iccvPaperID{****} % *** Enter the ICCV Paper ID here

\ificcvfinal\fi

\begin{document}

%%%%%%%%% TITLE
\title{Rethinking Local Perception in Lightweight Vision Transformer}

\author{Qihang Fan$^{1,2}$,  Huaibo Huang$^{2, 3}$,  Jiyang Guan$^{2,3}$,  Ran He$^{2,3}$\thanks{Corresponding author.}\\
$^1$Tsinghua University, China\\$^2$MAIS \& CRIPAC, Institute of Automation, Chinese Academy of Sciences, China\\$^3$School of Artificial Intelligence, University of Chinese Academy of Sciences, China\\
{\tt\small fqh19@mails.tsinghua.edu.cn, huaibo.huang@cripac.ia.ac.cn},\\
{\tt\small guanjiyang2020@ia.ac.cn, rhe@nlpr.ia.ac.cn}}

\maketitle
% Remove page # from the first page of camera-ready.
\ificcvfinal\thispagestyle{empty}\fi

%%%%%%%%% ABSTRACT
\begin{abstract}
   Vision Transformers (ViTs) have been shown to be effective in various vision tasks. However, resizing them to a mobile-friendly size leads to significant performance degradation. Therefore, developing lightweight vision transformers has become a crucial area of research. This paper introduces CloFormer, a lightweight vision transformer that leverages context-aware local enhancement. CloFormer explores the relationship between globally shared weights often used in vanilla convolutional operators and token-specific context-aware weights appearing in attention, then proposes an effective and straightforward module to capture high-frequency local information. In CloFormer, we introduce AttnConv, a convolution operator in attention's style. The proposed AttnConv uses shared weights to aggregate local information and deploys carefully designed context-aware weights to enhance local features. The combination of the AttnConv and vanilla attention which uses pooling to reduce FLOPs in CloFormer enables the model to perceive high-frequency and low-frequency information. Extensive experiments were conducted in image classification, object detection, and semantic segmentation, demonstrating the superiority of CloFormer. The code is available at \url{https://github.com/qhfan/CloFormer}
\end{abstract}

%%%%%%%%% BODY TEXT
\section{Introduction}

Recently, Vision Transformers (ViTs) have demonstrated superior performance in various vision tasks, such as visual recognition, object detection, and semantic segmentation. After the original ViT was introduced by Dosovitskiy~\cite{vit}, several variants~\cite{SwinTransformer, cvt, dat, pvt, uniformer, maxvit} have been proposed to reduce computational complexity. For instance, Swin-Transformer~\cite{SwinTransformer} conducts attention in local non-overlapping windows, while PVT~\cite{pvt} uses average pooling to merge tokens. However, these ViTs are unsuitable for mobile devices due to their big model size and high FLOPs. And their performance degrades dramatically when directly scaled down to a mobile-friendly size. Therefore, recent works~\cite{mobilevit, edgevit, LVT, mobileformer, parcnet} have focused on exploring lightweight vision transformers to make ViTs applicable to mobile devices. For example, MobileViT~\cite{mobilevit} investigates how to combine CNNs with transformers, while MobileFormer~\cite{mobileformer} fuses lightweight MobileNet~\cite{mobilenet} with transformers. Additionally, the recent EdgeViT~\cite{edgevit} proposes a local-global-local module to aggregate information. All of the above works aim to design mobile-friendly models with high performance, fewer parameters, and low FLOPs.

\begin{figure}[t]
    \centering
    \includegraphics[scale=0.54]{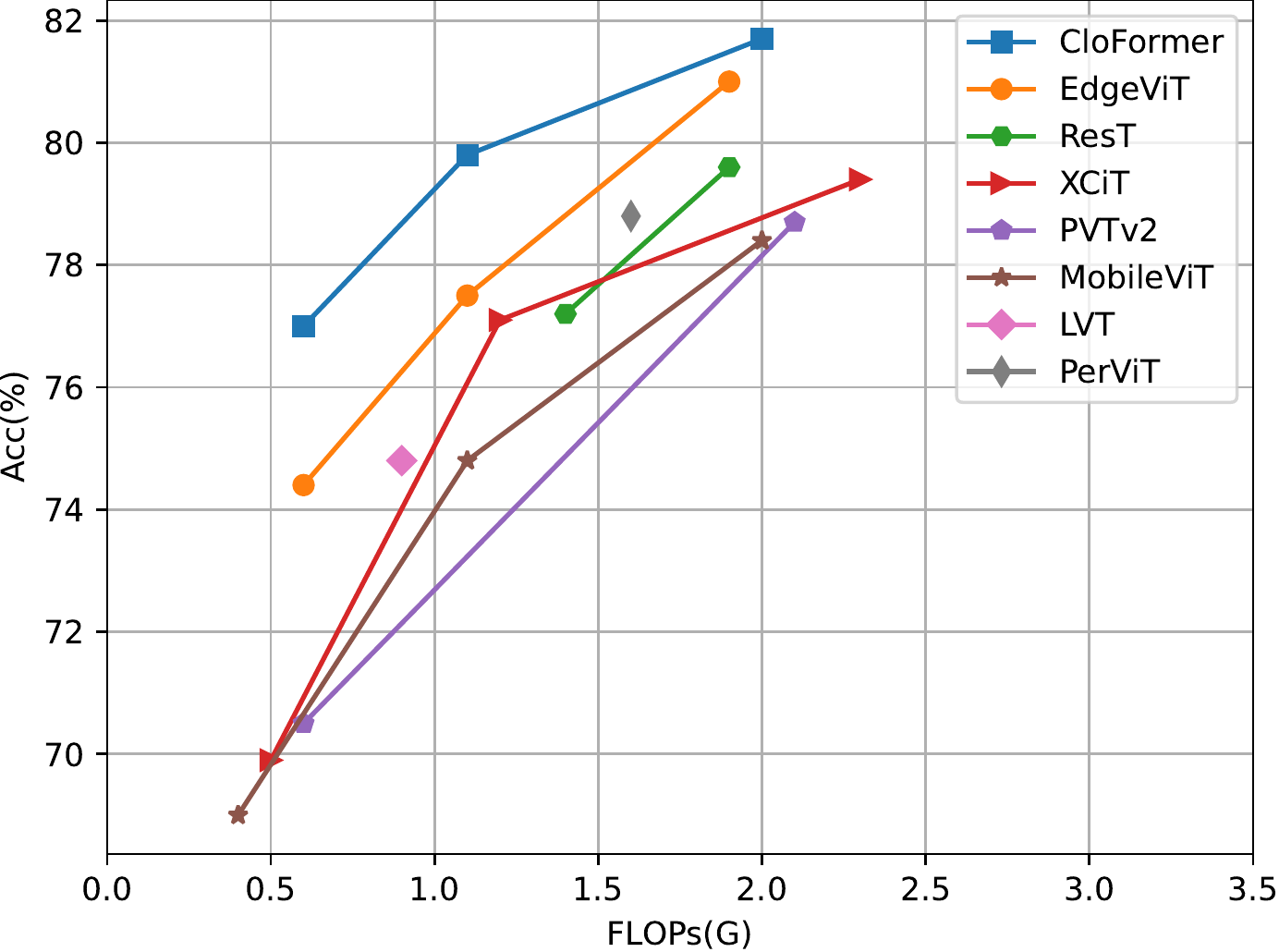}
    \caption{Comparision among high-performance models.}
    \label{fig:curve}
    \vspace{-0.15cm}
\end{figure}

However, in the existing lightweight models, most methods only focus on designing sparse attention to effectively process low-frequency global information while processing high-frequency local information using relatively simple methods. Specifically, most models, such as EdgeViT~\cite{edgevit} and MobileViT~\cite{mobilevit}, only use primitive convolution to extract local representations, and these methods only use globally shared weights that convolution has to deal with high-frequency local information. Other methods, such as LVT~\cite{LVT}, first unfold tokens into windows and then uses attention~\cite{SwinTransformer} within the windows to obtain high-frequency information. These methods only use context-aware weights that are specific to each token for local perception. Although the above lightweight models have achieved good results on multiple datasets, no method attempts to design a more effective way that leverages the advantages of shared and context-aware weights to process high-frequency local information. Shared weights-based methods, such as traditional convolutional neural networks, have the characteristics of translation equivariance. Different from them, context-aware weights-based methods such as LVT~\cite{LVT} and NAT~\cite{nat} have weights that can vary with input content. Both types of weights have their own advantages in local perception.

To simultaneously leverage the advantages of shared and context-aware weights, we introduce CloFormer, a lightweight vision transformer with context-aware local enhancement. CloFormer employs a two-branch structure. In the local branch, we introduce our carefully designed AttnConv, a simple and effective convolution operator in attention's style. The proposed AttnConv effectively fuses shared and context-aware weights to aggregate high-frequency local information. First, AttnConv utilizes depth-wise convolution (DWconv), which has shared weights to extract local representations. Next, it deploys context-aware weights to enhance local features. Unlike previous methods that generate context-aware weights through local self-attention~\cite{SwinTransformer, nat, LVT}, AttnConv generates context-aware weights using a gating mechanism, which introduces stronger nonlinearity than commonly used attention mechanisms. AttnConv applies convolution operators to $Q$ and $K$ to aggregate local information, then computes the Hardmard product of $Q$ and $K$, and performs a series of linear or nonlinear transformations on the results to generate context-aware weights within the range of $[-1, 1]$. It is noteworthy that AttnConv inherits the translation equivariance of convolution, as all its operators are based on convolution. 

As for the global branch, we use the vanilla attention but downsample the $K$ and $V$ to reduce FLOPs, which helps the model capture the low-frequency global information. In addition, a straightforward method is used to fuse the outputs of the local and global branches. This two-branch structure allows CloFormer to capture both high- and low-frequency information.

Extensive experiments on popular vision tasks, such as ImageNet classification~\cite{imagenet}, COCO object detection/instance segmentation~\cite{coco}, and ADE20K semantic segmentation~\cite{ade20k}, demonstrate the superiority of our CloFormer. As depicted in Fig.~\ref{fig:curve}, CloFormer achieves the best performance among models with similar FLOPs. Furthermore, in our ablation experiment, spectral analysis demonstrates that our approach, which combines a global and local branch, enables the perception of both high and low-frequency information. 

Our contributions are summarized as follows:
\begin{itemize}

\item In CloFormer, we introduce a convolution operator called AttnConv, which adopts the style of attention and fully leverages the benefits of shared and context-aware weights for local perception. Furthermore, it uses a new approach that incorporates stronger nonlinearity than vanilla local self-attention for generating context-aware weights.

\item In CloFormer, we utilize a two-branch architecture where one branch captures high-frequency information using AttnConv, while the other branch captures low-frequency information using vanilla attention with downsampling. The two-branch structure enables CloFormer to capture high-frequency and low-frequency information simultaneously.

\item Our extensive experiments on image classification, object detection and semantic segmentation demonstrate the superiority of CloFormer. Notably, CloFormer achieves an accuracy of 77.0\% on ImageNet1k with a mere 4.2M parameters and 0.6G FLOPs, outperforming other advanced models with comparable model sizes and FLOPs by a significant margin.
\end{itemize}

\begin{figure*}[t]
  \centering
   \includegraphics[scale=0.52]{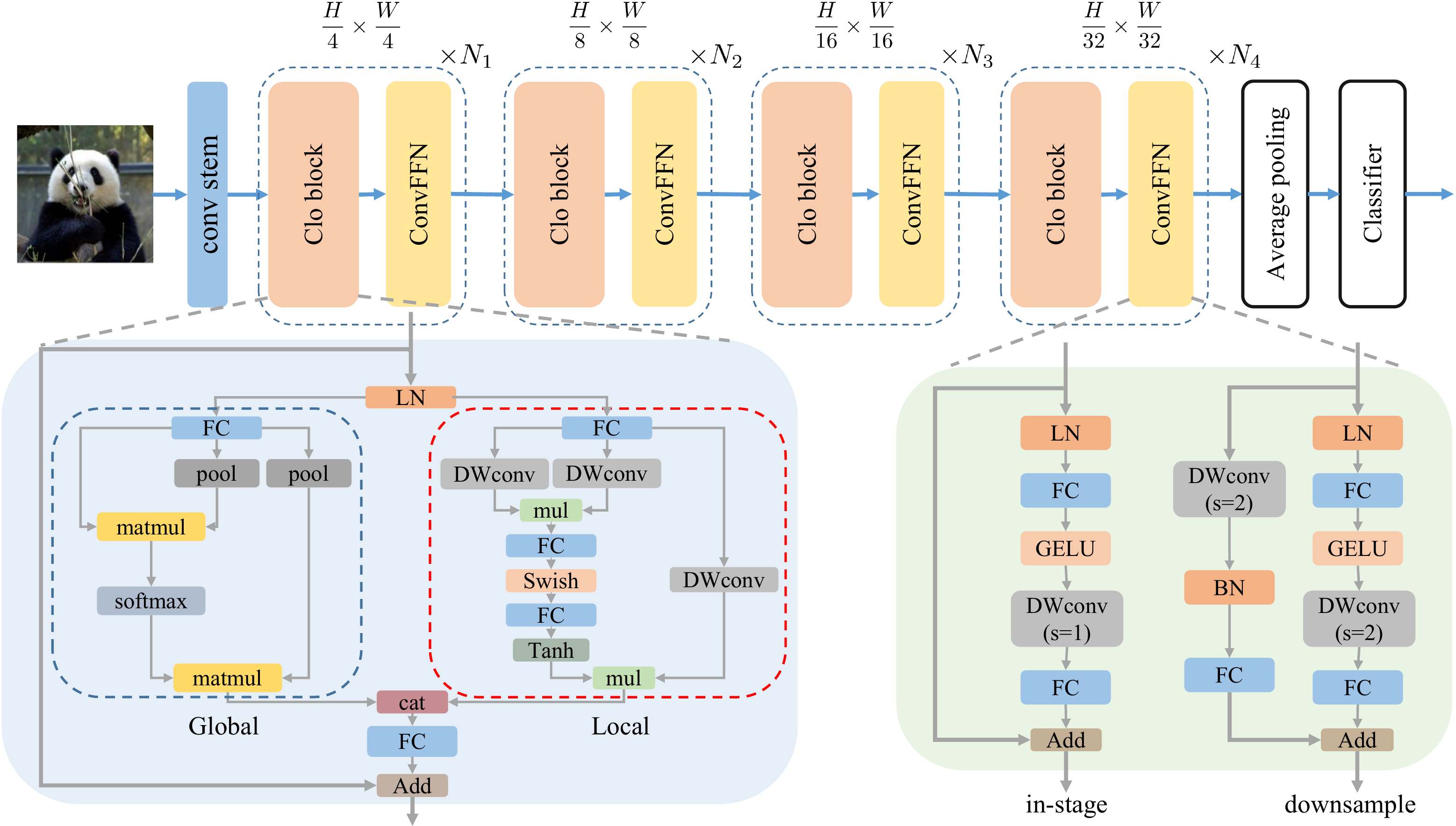}
   \caption{The illustration of CloFormer. CloFormer is constructed by connecting Clo block and ConvFFN in series. All depth-wise convolutions in the local branch have the stride of 1.}
   \label{fig:main}
   \vspace{-0.45cm}
\end{figure*}
\vspace{-0.25cm}
\section{Related Work}
\paragraph{Vision Transformers.}Since its introduction in recent years, ViT~\cite{vit} has quickly become a popular visual architecture due to its excellent global modeling capability. A series of works follow the original ViT and attain considerable improvement over it in terms of data efficiency~\cite{deit, cait} and architecture design~\cite{cvt, localvit, davit, cswin, qna, conformer, CPVT}. Among them, DeiT~\cite{deit} reduces ViTs' dependence on data through knowledge distillation, while others try to reduce its computational complexity~\cite{pvt, focal, litv1, litv2}. Most works reduce computational complexity by limiting the tokens that perform attention to a certain area~\cite{SwinTransformer, cswin, nat, LVT, twins}. The best-known one is Swin-Transformer~\cite{SwinTransformer}, which uses window self-attention to reduce FLOPs. Besides, some alternative approaches select tokens dynamically when performing attention~\cite{unkown, dynamicvit, evovit}. Specifically, in the forward pass, tokens containing little information are pruned or pooled, so the computational complexity is reduced. All the above works try to improve ViT's performance and make ViT flexible. 
\vspace{-0.35cm}
\paragraph{Lightweight Vision Transformers.}Most existing lightweight vision transformers (10M parameters and 2G FLOPs) are obtained by directly scaling down the large models~\cite{pvt, pervit, pvtv2, xcit, ortho, poolformer, maxvit}. However, when scaling down these models to a mobile-friendly size, their performance degrades dramatically. Therefore, in recent years, many works have focused on the design of lightweight transformers~\cite {edgevit, mobileformer, mobilevit, levit}. MobileViT~\cite{mobilevit} is a well-known lightweight transformer that connects convolution with attention and achieves high performance with few parameters. While MobileFormer~\cite{mobileformer} combines transformer and MobileNet~\cite{mobilenet} in parallel. The recently proposed EdgeViT~\cite{edgevit} uses a local-global-local module to aggregate information and get competitive results.
\vspace{-0.35cm}
\paragraph{Local Perception.}Among existing methods, two approaches are commonly used to aggregate local information. One is to employ convolution with shared weights, which is a fundamental component of convolutional neural networks and utilized in all traditional CNNs~\cite{vgg, resnet, huang2017densely, googlenet, nin}. The other approach involves using context-aware weights closely related to local self-attention~\cite{LVT, SwinTransformer, yuan2022volo, involution}. In LVT~\cite{LVT}, tokens are unfolded into windows, and attention is performed within each window. In Swin-Transformer~\cite{SwinTransformer}, tokens are directly divided into windows, and the similarities among tokens within a window are computed. These methods all use similarity scores as context-aware weights. However, unlike existing works, in CloFormer, we attempt to effectively leverage the advantages of shared and context-aware weights in local perception. Moreover, we utilize a method with stronger nonlinearity than attention to generate context-aware weights.

\vspace{-0.25cm}
\section{Method}

\subsection{Overall architecture}

As depicted in Fig.~\ref{fig:main}, CloFormer has four stages and includes the Clo block, ConvFFN, and convolution stem. We first pass the input image through the convolution stem to obtain tokens. The stem is comprised of four convolutions, each with strides of 2, 2, 1, and 1, respectively. Subsequently, the tokens undergo four stages of Clo block and ConvFFN to extract hierarchical features. Finally, we leverage global average pooling and a fully-connected layer to generate predictions.
\vspace{-0.35cm}
\paragraph{ConvFFN.}To incorporate local information into the FFN process, we replace the vanilla FFN with ConvFFN. The primary distinction between the ConvFFN and the commonly used FFN is that the ConvFFN employs a depth-wise convolution (DWconv) following the GELU activation, which enables ConvFFN to aggregate local information. Thanks to the DWconv, downsampling can be performed directly within ConvFFN without introducing the PatchMerge module. CloFormer employs two types of ConvFFN. The first is in-stage ConvFFN, which utilizes the skip connection directly. The other is the ConvFFN which connects two stages. In the skip connection of this type of ConvFFN, the DWconv and fully-connected layer are utilized to downsample and up-dimension the input, respectively.
\vspace{-0.35cm}
\paragraph{Clo block.}Clo blocks play a crucial role in CloFormer. Each block consists of a local branch and a global branch. In the global branch, as depicted in Fig.~\ref{fig:main}, we first downsample $K$ and $V$, and then perform the standard attention process on $Q$, $K$, and $V$ to extract low-frequency global information:
\begin{equation}
\label{eq:FC}
\centering
\begin{split}
    X_{global}=\bm{{\rm Attntion}}{\rm (}Q_g, \bm{{\rm Pool}}{\rm (}K_g{\rm )}, \bm{{\rm Pool}}{\rm (}V_g{\rm )}{\rm )}
\end{split}
\end{equation}
In addition to it, we utilize the proposed AttnConv as the local branch.

\subsection{AttnConv}

The pattern of the global branch effectively reduces the number of FLOPs required for attention and also results in a global receptive field. However, while it effectively captures low-frequency global information, it has insufficient capacity to process high-frequency local information~\cite{inceptionformer}. To address this limitation, we use AttnConv as a solution.

AttnConv is a crucial module that enables our model to achieve high performance. It incorporates some of the standard attention operations. Specifically, in AttnConv, we first apply a linear transformation to obtain the $Q$, $K$, and $V$, which is the same as the standard attention:
\begin{equation}
\label{eq:FC}
\centering
\begin{split}
    Q, K, V = \bm{{\rm FC}}{\rm (}X_{in}{\rm )}
\end{split}
\end{equation}
In Eq.~\ref{eq:FC}, $X_{in}$ is the input of the AttnConv. $\bm{{\rm FC}}$ denotes the fully-connected layer. After performing the linear transformation, we first conduct the local feature aggregation process with shared weights on $V$. Then, based on processed $V$ and $Q$, $K$, we perform the context-aware local enhancement.
\vspace{-0.65cm}
\paragraph{Local Feature Aggregation.}As shown in Fig.~\ref{fig:main}, for $V$, we use a simple depth-wise convolution (DWconv) to aggregate local information. The weights of DWconv are globally shared:
\begin{equation}
\label{eq:static_v}
    V_{s} = \bm{{\rm DWconv}}{\rm (}V{\rm )}
    \vspace{-0.3cm}
\end{equation}
\paragraph{Context-aware Local Enhancement.}After integrating local information for $V$ with shared weights, we combine $Q$ and $K$ to generate context-aware weights. It is worth noting that we use a different approach from local self-attention, which is more elaborate. Specifically, we first use two DWconvs to aggregate local information for $Q$ and $K$, respectively. Then, we calculate the Hadamard product of $Q$ and $K$ and perform a series of transformations on the result to obtain context-aware weights between $-1$ and $1$. Finally, we use the generated weights to enhance local features. The overall process can be summarized as follows:
\begin{equation}
\label{eq:attn}
\centering
\begin{split}
    &Q_l = \bm{{\rm DWconv}}{\rm (}Q{\rm )} \\
    &K_l = \bm{{\rm DWconv}}{\rm (}K{\rm )} \\
    &Attn_t = \bm{{\rm FC(Swish(FC(}}Q_l \odot K_l\bm{{\rm )))}}\\
    &Attn = \bm{{\rm Tanh}}{\rm (}\frac{Attn_t}{\sqrt{d}}{\rm )}\\
    &X_{local} = Attn \odot V_s\\
\end{split}
\end{equation}
In Eq.~\ref{eq:attn}, $d$ is the number of token's channels, and $\odot$ denotes the Hardmard product. Compared to vanilla attention, our method introduces stronger nonlinearity. Specifically, the only nonlinear operator in the context-aware weights generation process of local self-attention is $\bm{{\rm Softmax}}$. Nevertheless, in AttnConv, in addition to the $\bm{{\rm Tanh}}$, the $\bm{{\rm Swish}}$ is conducted. The stronger nonlinearity leads to higher-quality context-aware weights. 
\vspace{-0.35cm}
\paragraph{Fusion with Global Branch.}We use a straightforward method to fuse the output of the local branch with that of the global branch. Specifically, we concatenate the two outputs in the dimension of the channel. After that, a fully-connected layer is applied to the channel dimension:
\begin{equation}
\label{eq:fuse}
\centering
\begin{split}
    &X_{t} = \bm{{\rm Concat}}{\rm (}X_{local}, X_{global}{\rm )}\\
    &X_{out} = \bm{{\rm FC}}{\rm (}X_{t}{\rm )}\\
\end{split}
\end{equation}

\subsection{Different Ways of Local Perception}
\begin{figure}[t]
    \centering
    \includegraphics[scale=0.33]{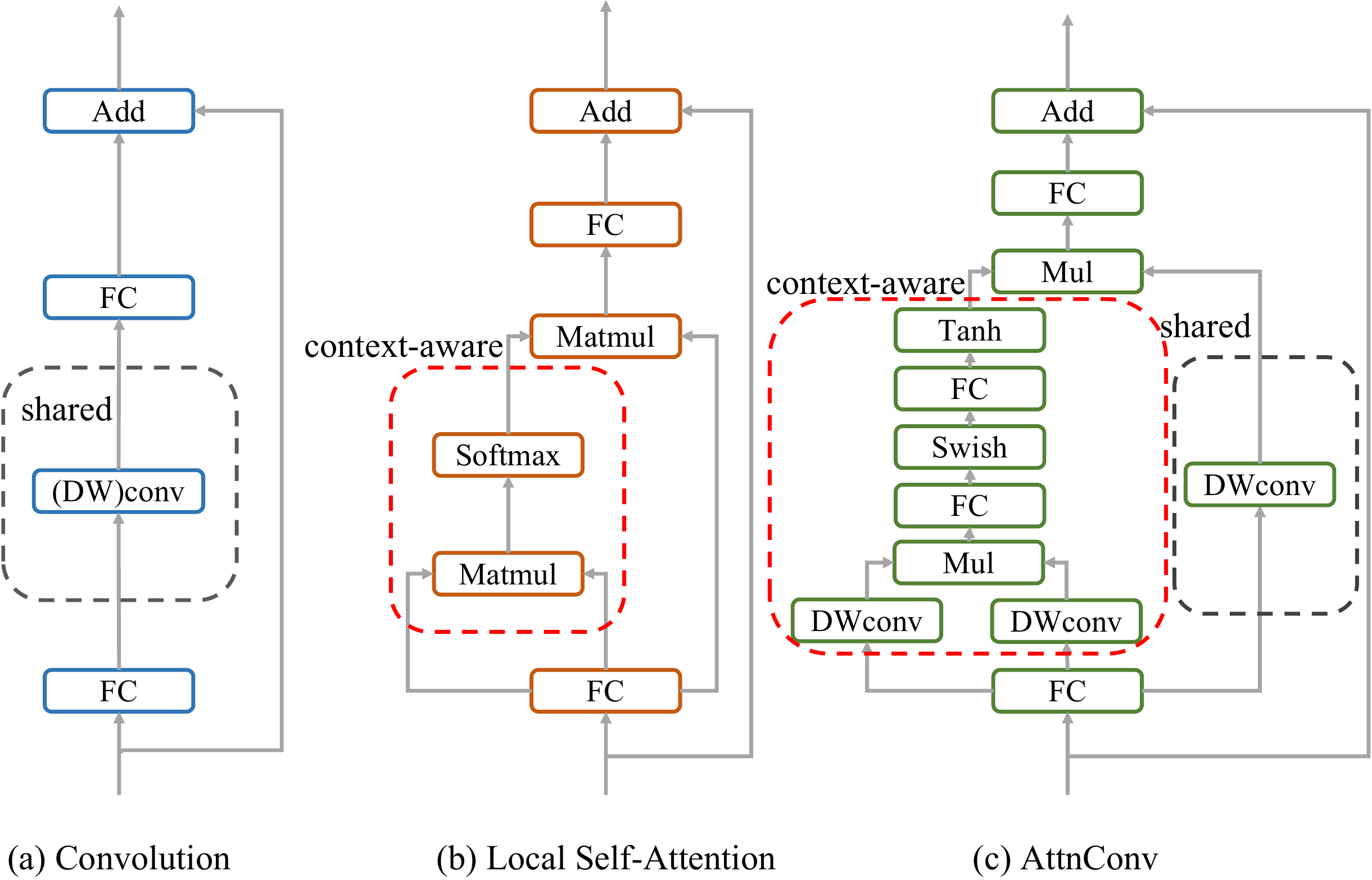}
    \caption{Comparision among different methods.}
    \label{fig:token}
    \vspace{-0.55cm}
\end{figure}
As shown in Fig.~\ref{fig:token}, we compare our AttnConv with traditional convolution and local self-attention. They all use residual structures. However, the operators used on their trunks are different.

\begin{table*}
    \centering
    \begin{tabular}{c|c c c c|c c}
    \hline
         Model & Blocks & Channels & Heads & Kernel Size& Params(M) & FLOPs(G)\\
         \hline
         CloFormer-XXS & [2, 2, 6, 2] & [32, 64, 128, 256] & [4, 4, 8, 16] & [3, 5, 7, 9] & 4.2 & 0.6\\
         CloFormer-XS & [2, 2, 6, 2] & [48, 96, 160, 352] & [3, 6, 10, 22] & [3, 5, 7, 9] & 7.2 & 1.1\\
         CloFormer-S & [2, 2, 6, 2] & [64, 128, 224, 448] & [4, 8, 14, 28] & [3, 5, 7, 9] & 12.3 & 2.0\\
    \hline
    \end{tabular}
    \caption{Architecture variants of CloFormer. The FLOPs are measured at
resolution $224\times224$. } 
    \label{tab:model}
    \vspace{-0.35cm}
\end{table*}

\vspace{-0.35cm}
\paragraph{Convolution.}The traditional residual block solely relies on convolution operators to gather high-frequency local information. As depicted in Fig.~\ref{fig:token}(a), for each token, the convolution operator performs a weighted sum on its neighboring tokens using the weights in the convolutional kernel. The weights in the kernel are globally shared and remain the same for different tokens.
\vspace{-0.35cm}
\paragraph{Local Self-Attention.}In contrast to convolution, local self-attention, as illustrated in Fig.~\ref{fig:token}(b), enables each token to gather information from its neighboring tokens by token-specific weights. This approach utilizes context-aware weights to extract high-frequency local representations, where tokens at different locations compute similarity scores with the neighboring tokens and gather information based on the similarity scores. 
\vspace{-0.35cm}
\paragraph{AttnConv.}We describe AttnConv in Fig.~\ref{fig:token}(c). After $Q$, $K$, and $V$ are obtained through linear transformation, we first use shared weights (DWconv) to conduct information aggregation on $V$. After that, we use the method that employs stronger nonlinearity than traditional attention to generate context-aware weights. Then we use these weights to enhance local features. In our AttnConv, we leverage both shared and context-aware weights.
\vspace{-0.35cm}
\paragraph{Advantages of AttnConv.}Compared to traditional convolution, the utilization of context-aware weights in AttnConv allows the model to better adapt to the input content during local perception. Compared to local self-attention, the introduction of shared weights enables our model to handle high-frequency information better, leading to improved performance. Besides, our approach for generating context-aware weights introduces stronger nonlinearity than local self-attention, which leads to better performance. It should be noted that all operations used in AttnConv are based on convolution, preserving the translation equivariance property of convolution.

\subsection{Implementation Details}
In Tab.~\ref{tab:model}, we present three variants of CloFormer based on the Clo block and ConvFFN. The ConvFFN expansion ratios are set to 4, and the kernel sizes are set to 5 for all ConvFFN layers. To capture high-frequency features in the early stages and low-frequency features in later stages, we gradually increase the kernel sizes used in AttnConv. Specifically, we set the kernel size of AttnConv to 3 in the first stage and 9 in the last stage.

\section{Experiment}

\begin{table*}[h]
    \centering
    \vspace{-7mm}
    \setlength{\tabcolsep}{1.5mm}
    \subfloat{
    \scalebox{0.8}{
    \begin{tabular}{c|c c|c|c c c}
    \hline
         Model & \makecell{Params\\(M)} & \makecell{FLOPs\\(G)} & \makecell{Top1\\(\%)}&\makecell{CPU\\(ms)} & \makecell{GPU\\(imgs/s)} &\makecell{Mem\\(GB)}\\
         \hline
         %\multirow{16}{*}{\makecell{XXS\\$4$M\\$0.6$G}} 
         %&XCiT-N12~\cite{xcit}  & 3.1 & 0.5 & 69.9 & 38.1 & 2582\\
         PVTv2-B0~\cite{pvtv2} &  3.4 & 0.6 & 70.5 & 44.5 & 1682 & 3.4\\
         MobileNetV1~\cite{mobilenet} &  4.2 & 0.6 & 70.6 & 29.8 & 3301 & 3.8\\
         T2T-ViT-7~\cite{t2t} & 4.3 & 1.1 & 71.7 & 42.8 & 1762 & 3.2\\
         QuadTree-B-b0~\cite{quadtree}  & 3.5 & 0.7 & 72.0 & --- & 885 & 4.4\\
         PiT-Ti~\cite{pit} &  4.9 & 0.7 & 73.0 & 31.6 & 2262 & 3.1\\
         %&Ortho-T~\cite{ortho} & 3.9 & 0.7 & 74.0 & --- & ---\\
         %&RegNetY-400MF &  5.3 & 0.4 & 74.1 & 52.6 & 3842\\
         EdgeViT-XXS~\cite{edgevit} & 4.1 & 0.6 & 74.4 & 42.1 & 1926 & 2.6\\
         MobileNetV2~\cite{mobilenetv2}$\times1.4$ &  6.1 & 0.6 & 74.7 & 34.5 & 2442 & 5.8\\
         MobileViT-XS~\cite{mobilevit} &  2.3 & 1.1 & 74.8 & 141.5 & 1367 & 10.0\\
         LVT~\cite{LVT} &  5.5 & 0.9 & 74.8 & 63.1 & 1265 & 3.7\\
         %&ViT-C~\cite{early} &  4.6 & 1.1 & 75.3 & \\
         VAN-B0~\cite{VAN} &  4.1 & 0.9 & 75.4 & 90.1 & 1662 & 5.3\\
         ShuffleNetV2$2\times$~\cite{shufflenetv2} & 5.5 & 0.6 & 75.4 & 36.2 & 3855 & 2.2\\
         tiny-MOAT-0~\cite{MOAT} & 3.4 & 0.8 & 75.5 & 61.1 & 2034 & 6.5\\
         RegNetY-800MF~\cite{regnety}  & 6.3 & 0.8 &76.3 & 65.8 & 2720 & 2.5 \\
         MobileFormer-214M~\cite{mobileformer} &  9.4 & 0.2 & 76.7 & 43.2 & 1744 & 3.7\\
         \rowcolor{yellow!60} CloFormer-XXS & 4.2 & 0.6 & 77.0 & 44.1 & 2425 & 3.4\\
         \hline
         %\multirow{10}{*}{\makecell{XS\\$7$M\\$1.0$G}}
         DeiT-Ti~\cite{deit} & 5.7 & 1.3 & 72.2 & 40.0 & 2256 & 2.7\\
         TNT-Tiny~\cite{tnt} &  6.1 & 1.4 & 73.9 & 78.2 & 545 & 5.4\\
         %&CPVT-Ti-GAP~\cite{CPVT} &  6 & 1.3 & 74.9 \\
         PVT-T~\cite{pvt} & 13.2 & 1.6 & 75.1 & 77.6 & 1214 & 4.8\\
         ViL-Tiny~\cite{ViL} & 6.7 & 1.4 & 76.3 & 71.2 & 857 & 4.7\\
         %&CeiT-T  & 6.4 & 1.2 & 76.4 & 75.1 & 842\\
         T2T-ViT-12~\cite{t2t}  & 6.9 & 1.9 & 76.5 & 59.8 & 1307 & 4.3\\
         XCiT-T12~\cite{xcit} & 6.7 & 1.2 & 77.1 & 62.5 & 1672 & 4.0\\
         Rest-lite~\cite{rest} &  10.5 & 1.4 & 77.2 & 71.3 & 1123 & 4.1\\
         %&Swin-1G~\cite{SwinTransformer}  & 7.3 & 1.0 & 77.3\\
         CoaT-Lite-T~\cite{coat}  & 5.7 & 1.6 & 77.5 & 84.5 & 1045 & 5.6\\
         EdgeViT-XS~\cite{edgevit}  & 6.7 & 1.1 & 77.5 & 62.6 & 1528 & 3.4\\
         \hline
    \end{tabular}}}
    \subfloat{
    \scalebox{0.8}{
    \begin{tabular}{c|c c|c|c c c}
    \hline
         Model & \makecell{Params\\(M)} & \makecell{FLOPs\\(G)} & \makecell{Top1\\(\%)}&\makecell{CPU\\(ms)} & \makecell{GPU\\(imgs/s)}&\makecell{Mem\\(GB)}\\
         \hline
         %\multirow{10}{*}{\makecell{XS\\$7$M\\$1.0$G}}
         %SiT-Ti w$/$o FRD & 15.9 & 1.0 & 77.7 & 91.4 & 1057 & 5.1\\
         RegNetY-1.6GF~\cite{regnety}  & 11.2 & 1.6 & 78.0 & 111.3 & 1241 & 4.7\\
         PiT-XS~\cite{pit}  & 10.6 & 1.4 & 78.1 & 51.4 & 1682 & 3.7\\
         MPViT-T~\cite{mpvit} & 5.8 & 1.6 & 78.2 & 111.3 & 737 & 8.1\\
         %&DFvT-S~\cite{dfvit}  & 11.2 & 0.8 & 78.3 & \\
         tiny-MOAT-1~\cite{MOAT} & 5.1 & 1.2 & 78.3 & 80.7 & 1529 & 8.0\\
         MobileViT-S~\cite{mobilevit} & 5.6 & 2.0 & 78.4 & 185.4 & 898 & 12.1\\
         ParC-Net-S~\cite{parcnet}  & 5.0 & 1.7 & 78.6 & 145.3 & 1321 & 9.1\\
         PerViT-T~\cite{pervit}  & 7.6 & 1.6 & 78.8 & 153.4 & 1402 & 4.1\\
         EfficientNet-B1~\cite{efficientnet} & 7.8 & 0.7 & 79.4 & 60.3 & 1395 & 8.6 \\
         %&FAN-T-ViT~\cite{FAN} & $224^2$ & 7.3 & 1.3 & 79.2 \\
         
         MobileFormer-508m~\cite{mobileformer} & 14.0 & 0.5 & 79.3 & 64.2 & 1654 & 6.4\\
         \rowcolor{yellow!60} CloFormer-XS  & 7.2 & 1.1 & 79.8 & 62.4 & 1676 & 4.7\\
         \hline
         %\multirow{14}{*}{\makecell{S\\$12$M\\$2.0$G}} 
         %& PoolFormer-S12 & 11.9 & 2.0 & 77.2 & 63.3 & 1722\\
         PVTv2-B1~\cite{pvt}  & 13.1 & 2.1 & 78.7 & 103.6 & 1007 & 6.3\\
         %CoaT-Lite-Mi & 11 & 2.0 & 79.1 & 95.2 & 963 & 6.3\\
         %&EfficientFormer-L1  & 12.3 & 1.3 & 79.2 & 65.3 & 1966\\
         XCiT-T24~\cite{xcit}  & 12.1 & 2.3 & 79.4 & 119.1 & 1194 & 7.0\\
         %&LeViT-384$\dag$~\cite{levit} & $224^2$ & \textbf{39.1} & 2.4 & 79.5\\
         %&ResT-Small &  13.7 & 1.9 & 79.6 & 90.3 & 918 & 5.2\\
         QuadTree-B-b1~\cite{quadtree}  & 13.6 & 2.3 & 80.0 & ---& 543 & 7.9\\
         MPViT-XS~\cite{mpvit}  & 10.5 & 2.9 & 80.9 & 153.6 & 597 & 10.5\\
         PiT-S~\cite{pit} & 23.5 & 2.9 & 80.9 & 93.6 & 1042 & 5.7\\
         EdgeViT-S~\cite{edgevit}  & 11.1 & 1.9 & 81.0 & 96.8 & 1049 & 5.1 \\
         tiny-MOAT-2~\cite{MOAT} & 9.8 & 2.3 & 81.0 & 122.1 & 1047 & 11.0\\
         RegNetY-4GF~\cite{regnety} & 21.0 & 4.0 & 81.5 & 122.5 & 776 & 6.8 \\
         EfficientNet-B3~\cite{efficientnet} & 12.0 & 1.8 & 81.6 & 124.2 & 634 & 18.7\\
         VAN-B1~\cite{VAN} &  13.9 & 2.5 & 81.1 & 155.2 & 995 & 7.6 \\
         PVT-S~\cite{pvt} & 24.5 & 3.8 & 79.8 & 136.3 & 677 & 8.1 \\
         DeiT-S~\cite{deit} & 22.1 & 4.6 & 79.9 & 113.6 & 899 & 5.5\\
         Swin-T~\cite{SwinTransformer} & 28.3 & 4.5 & 81.3 & 184.3 & 664 & 7.7\\
         \rowcolor{yellow!60} CloFormer-S & 12.3 & 2.0 & 81.6 & 93.1 & 1186 & 6.3\\
         \hline
    \end{tabular}}}
  
    \caption{ Comparison on ImageNet-1K classification. We measure the latency of models on Intel Core i9 with batch size of 1, and their throughput/memory peak on V100 32G with batch size of 64. The lower the latency on the CPU and the higher the throughput on the GPU, the faster inference speed of the model. 
    }
    \label{tab:imagenet}
\end{table*}

\begin{table*}[h]
    \centering
    \scalebox{0.84}{
    \begin{tabular}{c |c |c c c |c c c |c |c c c |c c c}
    \hline
        \multirow{2}{*}{Backbone} & \multicolumn{7}{c|}{RetinaNet $1\times$} &\multicolumn{7}{c}{Mask R-CNN $1\times$}\\
        \cline{2-15}
        & Par(M) & $AP$ & $AP_{50}$ & $AP_{75}$ & $AP_S$ & $AP_M$ & $AP_L$ & Par(M) & $AP^b$ & $AP^b_{50}$ & $AP^b_{75}$ & $AP^m$ & $AP^m_{50}$ & $AP^m_{75}$\\
        \hline
        PVTv2-B0~\cite{pvtv2} & 13.0 & 37.2 & 57.2 & 39.5 & 23.1 & 40.4 & 49.7 & 23.5 & 38.2 & 60.5 & 40.7 & 36.2 & 57.8 & 38.6 \\
        QuadTree-B-b0~\cite{quadtree} & 13.1 & 38.4 & 58.7 & 41.1 & 22.5 & 41.7 & 51.6 & 23.7 & 38.8 & 60.7 & 42.1 & 36.5 & 58.0 & 39.1 \\
        EdgeViT-XXS~\cite{edgevit} & 13.1 & 38.7 & 59.0 & 41.0 & 22.4 & 42.0 & 51.6 & 23.8 & 39.9 & 62.0 & 43.1 & 36.9 & 59.0 & 39.4 \\
        \textbf{CloFormer-XXS} & 13.4 & \textbf{39.6} & \textbf{60.3} & \textbf{42.1} & \textbf{24.2} & \textbf{43.1} & \textbf{53.2} & 23.9 & \textbf{40.3} & \textbf{62.6} & \textbf{43.4} & \textbf{37.3} & \textbf{59.6} & \textbf{39.5} \\
        \hline
        MobileFormer-294~\cite{mobileformer} & 16.1 & 36.6 & 56.6 & 38.6 & 21.9 & 39.5 & 47.9 & - & - & - & - & - & - & -\\
        DFvT-S~\cite{dfvit} & - & - & - & - & - & - & - & 32 & 39.2 & 62.2 & 42.4 & 36.3 & 58.9 & 38.6 \\ 
        EdgeViT-XS~\cite{edgevit} & 16.3 & 40.6 & 61.3 & 43.3 & 25.2 & 43.9 & 54.6 & 26.5 & 41.4 & 63.7 & 45.0 & 38.3 & 60.9 & 41.3\\
        ViL-Tiny~\cite{ViL} & 16.6 & 40.8 & 61.3 & 43.6 & 26.7 & 44.9 & 53.6 & 26.9 & 41.4 & 63.5 & 45.0 & 38.1 & 60.3 & 40.8 \\
        MPViT-T~\cite{mpvit} & 17 & 41.8 & 62.7 & 44.6 & \textbf{27.2} & 45.1 & 54.2 & 28& 42.2 & 64.2 & 45.8 & 39.0 & 61.4 & 41.8\\
        \textbf{CloFormer-XS} & 16.6 & \textbf{42.2} & \textbf{63.2} & \textbf{44.9} & 26.6 & \textbf{45.7} & \textbf{56.4} & 26.9 & \textbf{42.9} & \textbf{65.1} & \textbf{47.1} & \textbf{39.2} & \textbf{61.6} & \textbf{42.4} \\ 
        \hline
        ResNet18~\cite{resnet} & 21.3 & 31.8 & 49.6 & 33.6 & 16.3 & 34.3 & 43.2 & 31.2 & 34.0 & 54.0 & 36.7 & 31.2 & 51.0 & 32.7 \\
        PoolFormer-S12~\cite{poolformer} & 21.7 & 36.2 & 56.2 & 38.2 & 20.8 & 39.1 & 48.0 & 31.6 & 37.3 & 59.0 & 40.1 & 34.6 & 55.8 & 36.9 \\
        PVTv1-Tiny~\cite{pvt} & 23.0 & 36.7 & 56.9 & 38.9 & 22.6 & 38.8 & 50.0 & 32.9 & 36.7 & 59.2 & 39.3 & 35.1 & 56.7 & 37.3\\
        EfficientFormer-L1~\cite{efficientformer} & - & - & - & - & - & - & - & 31.9 & 37.9 & 60.3 & 41.0 & 35.4 & 57.3 & 37.3 \\
        ResT-Small~\cite{rest} & 23.4 & 40.3 & 61.3 & 42.7 & 25.7 & 43.7 & 51.2 & 33.3 & 39.6 & 62.9 & 42.3 & 37.2 & 59.8 & 39.7 \\
        PVTv2-B1~\cite{pvtv2} & 23.8 & 41.2 & 61.9 & 43.9 & 25.4 & 44.5 & 54.3 & 33.7 & 41.8 & 64.3 & 45.9 & 38.8 & 61.2 & 41.6\\
        VAN-B1~\cite{VAN} & 23.6 & 42.3 & 63.1 & 45.1 & 26.1 & 46.2 & 54.1 & 33.5 & 42.6 & 64.2 & 46.7 & 38.9 & 61.2 & 41.7 \\
        QuadTree-B-b1~\cite{quadtree} & 23.8 & 42.6 & 63.6 & 45.3 & 26.8 & 46.1 & 57.2 & 33.6 & 43.5 & 65.6 & 47.6 & 40.1 & 62.6 & 43.3\\
        MPViT-XS~\cite{mpvit} & 20 & 43.8 & 65.0 & 47.1 & \textbf{28.1} & 47.6 & 56.5 & 30 & 44.2 & 66.7 & 48.4 & 40.4 & 63.4 & 43.4\\
        EdgeViT-S\cite{edgevit} & 22.6 & 43.4 & 64.9 & 46.5 & 26.9 & 47.5 & 58.1 & 32.8 & 44.8 & 67.4 & 48.9 & 41.0 & 64.2 & 43.8\\
        \textbf{CloFormer-S} & 22.0 & \textbf{43.9} & \textbf{65.1} & \textbf{47.2} & 27.2 & \textbf{47.6} & \textbf{58.2} & 32.0 & \textbf{45.1} & \textbf{67.8} & \textbf{49.1} & \textbf{41.4} & \textbf{64.7} & \textbf{44.2} \\
        \hline
    \end{tabular}}
    \caption{Comparison to other backbones using RetinaNet and Mask-RCNN on COCO val2017 object detection and instance segmentation. "Par" is the number of parameters.}
    \label{tab:coco}
    \vspace{-0.45cm}
\end{table*}

We conduct experiments on a variety of vision tasks, including image classification on ImageNet1K~\cite{imagenet}, object detection and instance segmentation on COCO 2017~\cite{coco}, and semantic segmentation on ADE20K~\cite{ade20k}. We pre-train CloFormer on ImageNet1K for visual recognition task and evaluate the generalization ability of CloFormer on COCO 2017 and ADE20K for dense prediction. We also do ablation studies to validate the importance of critical components in CloFormer. Details about experimental settings and additional ablation studies can be found in \textcolor{red}{Appendix}.

\subsection{ImageNet1K Classification}

\paragraph{Experimental Settings.}We benchmark CloFormer on ImageNet1K classification~\cite{imagenet} and follow the same training strategy in DeiT~\cite{deit}. We use the same data augmentation and regularization as DeiT, except for repeated augmentation and EMA. All our models are trained from scratch for 300 epochs, and we set the input size to $224\times224$. As for the optimizer, we use AdamW~\cite{adamw} with a cosine decay learning rate scheduler. Following the DeiT, we set the initial learning rate, weight decay, and total batch size to 0.001, 0.05, and 1024, respectively. The maximum rates of stochastic depth are set to 0/0.06/0.06 from XXS to S. For a fair comparison, we do not use other supervision methods such as distillation~\cite{deit}. 
\vspace{-0.35cm}
\paragraph{Results.}We report the ImageNet1K classification results in Tab.~\ref{tab:imagenet}. The results demonstrate that our models perform better than previous models when they have similar model sizes and FLOPs. Specifically, the CloFormer-XXS achieves 77.0\% Top-1 accuracy with only 4.2M parameters and 0.6G FLOPs, which surpasses ShuffleNetV2 $2\times$, MobileViT-XS, and EdgeViT-XXS by 1.6\%, 2.2\% and 2.6\%, respectively. Moreover, CloFormer-XXS takes about half as many FLOPs as MobileViT-XS. Our CloFormer-XS performs very similarly to QuadTree-B-b1 but requires only 53\% of QuadTree-B-b1's parameters and 48\% of its FLOPs. As for models with about 12M parameters and 2.0GFLOPs, our CloFormer-S surpasses VAN-B1 by 0.5\% but uses fewer parameters and lower FLOPs. All above results verify CloFormer's superiority.

\begin{figure*}[!htbp]
    \centering
    \includegraphics[scale=0.54]{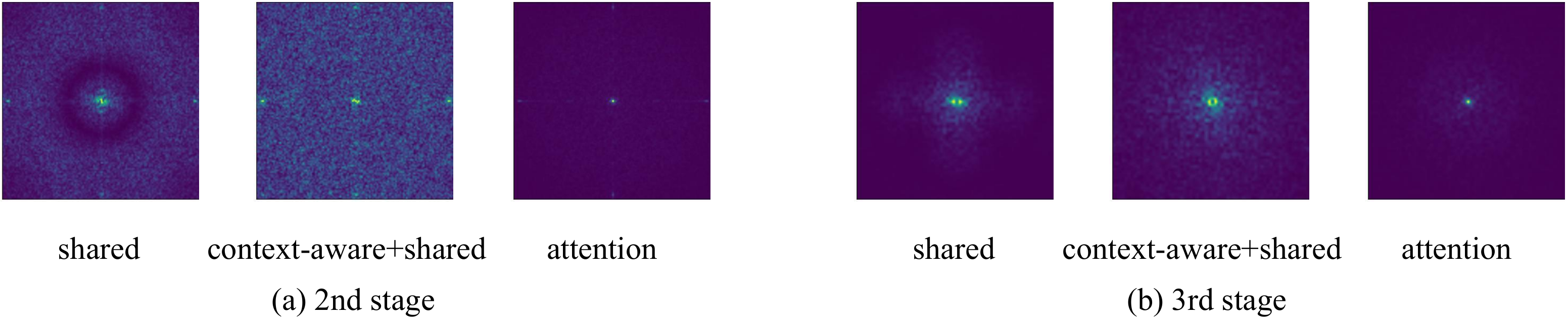}
    \caption{Spectral analysis of CloFormer.}
    \label{fig:freq}
    \vspace{-0.25cm}
\end{figure*}

\subsection{COCO Object Detection}

\paragraph{Experimental Settings.}We conduct object detection and instance segmentation on COCO 2017 dataset~\cite{coco}. Our experimental settings are following PVTv2~\cite{pvtv2}. We use the RetinaNet~\cite{retinanet} for object detection, and Mask R-CNN~\cite{maskrcnn} with the FPN~\cite{fpn} for instance segmentation. During training, images are resized to have a shorter side of 800 pixels. For the two tasks, we use the backbones pretrained on ImageNet1K, and we finetune them on the COCO training set with the batch size of 16 and the AdamW optimizer whose initial learning rate is set to $1\times10^{-4}$. Our implementation of coco object detection is based on MMDetection~\cite{mmdetection}. 
\vspace{-0.35cm}
\paragraph{Results.}Tab.~\ref{tab:coco} shows the results with RetinaNet and Mask R-CNN in $1\times$ schedule. The CloFormer achieves the best performance almost in all comparisons. As for RetinaNet, our CloFormer-XXS surpasses the recent EdgeViT by \textbf{+0.9} box AP, and CloFormer-XS outperforms MPViT-T by \textbf{+0.4} box AP. When we use Mask R-CNN, CloFormer-XXS surpasses EdgeViT-XXS by \textbf{+0.4} box AP and \textbf{+0.4} mask AP.

\subsection{ADE20K Semantic Segmentation}

\paragraph{Experimental Settings.}We conducted semantic segmentation on the ADE20K dataset~\cite{ade20k} with the Semantic FPN~\cite{semanticfpn}. The pretrained CloFormer is incorporated into the Semantic FPN. Following EdgeViT~\cite{edgevit}, during training, we create $512\times512$ random crops of the images, and during inference, we resize the images to have a shorter side of 512 pixels. We finetune the model for 80K iterations on ADE20K training set with AdamW, which uses an initial learning rate of $1\times10^{-4}$. The batch size is set to 16, and we report the mean Intersection over Union (mIoU) score on the validation set.
\vspace{-0.35cm}
\paragraph{Results.}In Tab.~\ref{tab:ade20k}, we compare our CloFormer to previous ViTs and CNNs in the framework of Semantic FPN~\cite{semanticfpn}. CloFormer achieves the best performance among these models. Specifically, our CloFormer-XXS outperforms EdgeViT-XXS by \textbf{+0.7}, and CloFormer-XS surpasses EdgeViT-XS by \textbf{+1.0} in mIoU.

\begin{table}[h]
    \centering
    \setlength{\tabcolsep}{1.1mm}
    \begin{tabular}{c|c|c|c}
    \hline
         \multirow{2}{*}{Backbone} & \multicolumn{3}{c}{Semantic FPN}\\
         \cline{2-4}
         & Params(M) & FLOPs(G) & mIoU(\%)\\
         \hline
         PVTv2-B0~\cite{pvtv2} & 7.6 & 25.0 & 37.2 \\
         VAN-B0~\cite{VAN} & 8 & 26 & 38.5\\
         EdgeViT-XXS~\cite{edgevit} & 7.9 & 24.4 & 39.7 \\
         CloFormer-XXS & 8.1 & 25.2 & \textbf{40.4} \\
         \hline
         EdgeViT-XS~\cite{edgevit} & 10.6 & 27.7 & 41.4 \\
         CloFormer-XS & 11.1 & 28.2 & \textbf{42.4}\\
         \hline
         ResNet18~\cite{resnet} & 15.5 & 32.2 & 32.9 \\
         PVTv1-Tiny~\cite{pvt} & 17.0 & 33.2 & 35.7 \\
         PoolFormer~\cite{poolformer} & 16.2 & 31.0 & 37.2 \\
         PVTv2-B1~\cite{pvtv2} & 17.8 & 34.2 & 42.5 \\
         VAN-B1~\cite{VAN} & 18 & 35 & 42.9 \\
         CloFormer-S & 16.2 & 34.6 & \textbf{44.2}\\
         \hline
    \end{tabular}
    \caption{Semantic segmentation results on the validation set of ADE20K.}
    \label{tab:ade20k}
    \vspace{-0.45cm}
\end{table}

\subsection{Ablation study}
In this section, we decouple the Clo block into multiple modules, and conduct detailed ablation studies. We verify the role of each part of AttnConv on ImageNet classification, COCO object detection, and ADE20K semantic segmentation.

\vspace{-0.35cm}
\paragraph{Spectral Analysis.}For spectrum analysis, we select two feature maps from the second and third stages of CloFormer. In Fig.~\ref{fig:freq}, we compare three feature maps: one that uses only shared weights (part of AttnConv), another that employs context-aware+shared weights (the full AttnConv), and a third that only utilizes the global branch. Our analysis indicates that the full AttnConv captures a greater number of high-frequency features as compared to shared weights only, while the attention branch is particularly effective at aggregating low-frequency features.

\begin{table*}[t]
    \centering
    \begin{tabular}{c|c c c|c c|c}
    \hline
         & \multicolumn{3}{c|}{ImageNet1K} & \multicolumn{2}{c|}{COCO} & ADE20K \\
         Model & Params(M) & FLOPs(G) & Top-1(\%) & $AP^b$ & $AP^m$ & mIoU\\
    \hline
         only global branch ($3\times3$) & 3.6 & 0.52 & 74.1 & 36.0 & 34.2 & 36.7\\
         $3\times3\xrightarrow{}5\times5$ & 4.0 & 0.58 & 74.7 & 37.0 & 34.8 & 37.2\\
         +shared weights& 3.9 & 0.55 & 75.2 & 38.4 & 35.3 & 38.6\\
         +$Q\odot K$ & 4.0 & 0.59 & 72.3 & --- & --- & --- \\
         +$\bm{{\rm Tanh}}$ & 4.0 & 0.59 & 76.0 & 39.6 & 36.3 & 39.2\\
         +$\bm{{\rm DWconv}}$ & 4.1 & 0.60 & 76.4 & 40.1 & 36.7 & 39.9\\
         +$\bm{{\rm FC}}$ & 4.2 & 0.62 & 76.5 & 40.1 & 36.9 & 39.9 \\
         +$\bm{{\rm Swish}}$ (full AttnConv) & 4.2 & 0.62 & \textbf{77.0} & \textbf{40.3} & \textbf{37.3} & \textbf{40.4}\\
         \hline
         only global branch & 4.0 & 0.58 & 74.7 & 37.0 & 34.8 & 37.2 \\
         only local branch & 4.7 & 0.69 & 76.7 & 39.9 & 36.8 & 40.0\\
         global branch+shared weights & 3.9 & 0.55 & 75.2 & 38.4 & 35.3 & 38.6\\
         global branch+context-aware weights & 4.1 & 0.62 & 76.4 & 39.7 & 36.7 & 39.6\\
         CloFormer & 4.2 & 0.62 & \textbf{77.0} & \textbf{40.3} & \textbf{37.3} & \textbf{40.4} \\
         \hline
    \end{tabular}
    \caption{Ablation of CloFormer.}
    \label{tab:abattn}
    \vspace{-0.1cm}
\end{table*}

\begin{table*}[h]
    \centering

    \begin{tabular}{c|c c c|c c|c}
    \hline
         & \multicolumn{3}{c|}{ImageNet1K} & \multicolumn{2}{c|}{COCO} & ADE20K \\
         Model & Params(M) & FLOPs(G) & Top-1(\%) & $AP^b$ & $AP^m$ & mIoU\\
    \hline
         shared weights & 3.9 & 0.55 & 75.2 & 38.4 & 35.3 & 38.6\\
         %context-aware (attention) & 4.0 & 0.60 & & & & \\
         %context-aware (AttnConv) & 4.1 & 0.62 & & & & \\
         \hline
         context-aware (attention)+shared weights& 4.0 & 0.61 & 75.8 & 39.2 & 36.1 & 39.5\\
         \hline
         AttnConv & 4.2 & 0.62 & \textbf{77.0} & \textbf{40.3} & \textbf{37.3} & \textbf{40.4}\\
         \hline
    \end{tabular}

    \caption{Comparison among shared weighs, context-aware (attention)+shared weights, and AttnConv.}
    \label{tab:3comp}
    \vspace{-0.25cm}
\end{table*}
\vspace{-0.35cm}
\paragraph{ConvFFN.}We first conduct experiments using a CloFormer model containing only the global branch. Moreover, we increase the kernel size of the DWconv in ConvFFN from 3 to 5. As shown in Tab.~\ref{tab:abattn}, the model performs better with a larger kernel size. With the addition of ConvFFN, whose kernel size is 5, the model's performance is comparable to that of LVT~\cite{LVT}. The result demonstrates that ConvFFN's DWconv has already captured some high-frequency local information.
\vspace{-0.35cm}
\paragraph{Shared Weights.}Tab.~\ref{tab:abattn} demonstrates that despite ConvFFN already capturing high-frequency local features, the inclusion of the local branch in addition to the global branch further improves model performance, even when only shared weights are utilized to aggregate high-frequency information. Specifically, the model's performance improves by 0.5\%, while the number of parameters decreases by 0.1M. These results suggest that the integration of high- and low-frequency information can provide additional benefits to the model.
\vspace{-0.35cm}
\paragraph{Directly use $Q$ and $K$.}We introduce context-aware weights after adding shared weights. Initially, we directly use the Hardmard product of $Q$ and $K$ as the context-aware weights. However, we observe a significant slowdown in the model's convergence rate when nothing is done to the Hardmard product of $Q$ and $K$. In fact, under these circumstances, the model's accuracy decreases by 2.9\% when compared to using only shared weights.
\vspace{-0.35cm}
\paragraph{Add Tanh.}Continuing from the previous step, we introduce context-aware weights that are activated using the $\bm{{\rm Tanh}}$. We then use the resulting weights to enhance local features. The impact of this approach is evident from Tab.~\ref{tab:abattn}, where the addition of $\bm{{\rm Tanh}}$ leads to a significant improvement in the convergence rate of the model, with a significantly higher Top1-acc compared to the original case after 300 epochs of training. This improved performance can be attributed to the weight normalization provided by $\bm{{\rm Tanh}}$, which constrains the weights between -1 and 1.
\vspace{-0.35cm}
\paragraph{Add DWconv.}Before calculating the Hadamard product of $Q$ and $K$, we incorporate two DWconvs to aggregate the local information for both $Q$ and $K$. The impact of this step is apparent from Tab.~\ref{tab:abattn}, where we observe improved model performance. This underscores the importance of incorporating local information in the context-aware weights generation process. By adding high-frequency local information, the model can generate high-quality context-aware weights, ultimately resulting in better performance. In fact, this step has brought about an increase of 0.4\% in the classification accuracy of our model.
\vspace{-0.35cm}
\paragraph{Effect of Nonlinearity.}We extend the process of generating context-aware weights by including two fully-connected layers in addition to $\bm{{\rm Tanh}}$ and $\bm{{\rm DWconv}}$. However, the addition of the linear transformation alone, as demonstrated in Tab.~\ref{tab:abattn}, does not provide significant improvement to the model's performance. Then, we introduce the $\bm{{\rm Swish}}$ function between the two fully-connected layers to enhance the nonlinearity of the weight generation process. Our experiments, as shown in Tab.~\ref{tab:abattn}, demonstrate that the inclusion of $\bm{{\rm Swish}}$ has led to a significant improvement (0.5\%) in the model's performance. More experiments can be found in \textcolor{red}{Appendix}.
\begin{figure}[h]
\vspace{-0.2cm}
    \centering
    \includegraphics[scale=0.30]{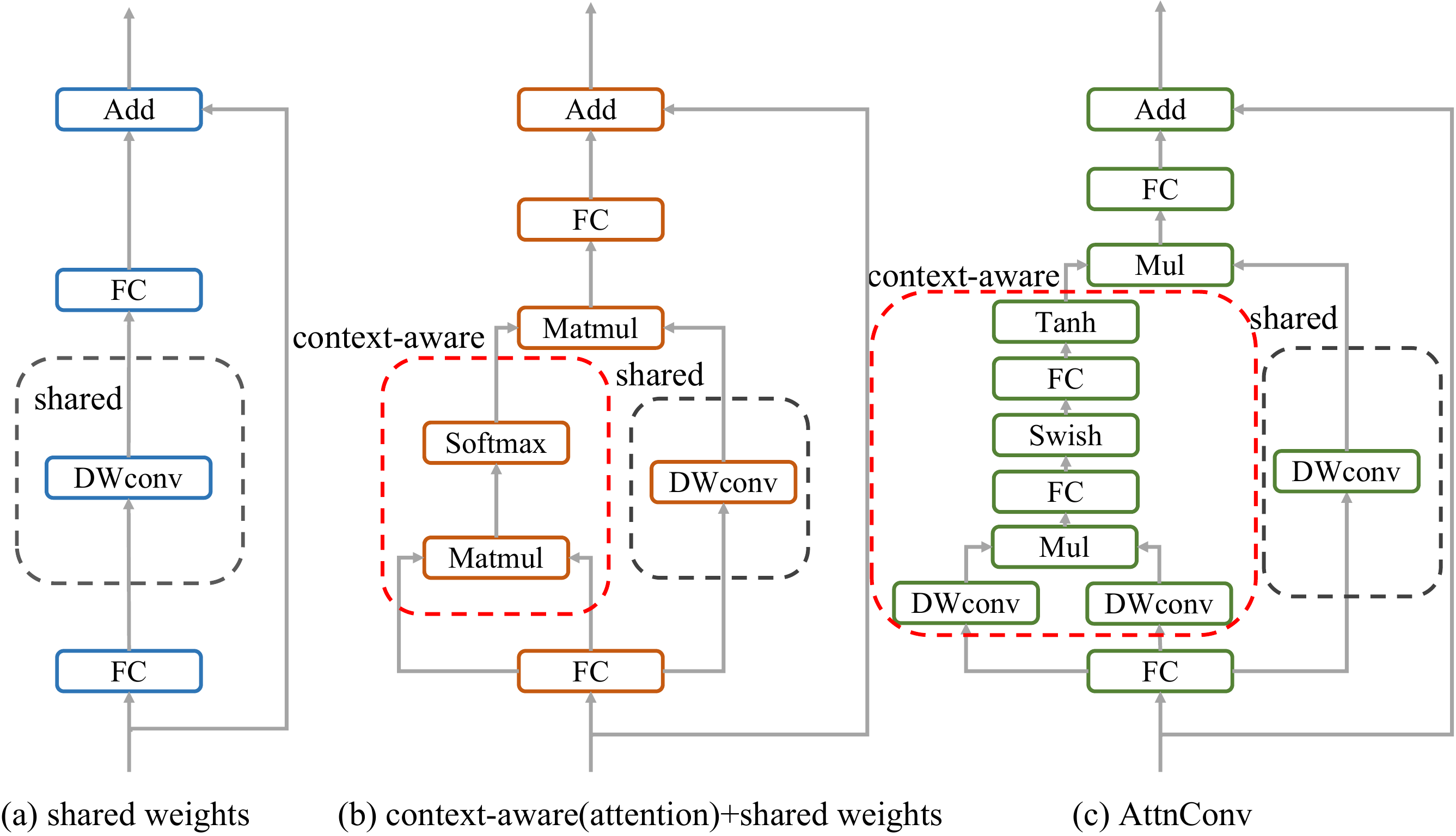}
    \caption{Transformation from convolution to AttnConv.}
    \label{fig:comp}
    \vspace{-0.45cm}
\end{figure}
\vspace{-0.35cm}
\paragraph{Local\&\&Global Branch.}As shown in Tab.~\ref{tab:abattn}, we compare the model's performance of using only the local branch, only the global branch, and the complete CloFormer. The complete CloFormer outperforms using only the local branch by 0.3\% and using only the global branch by 2.3\%.
\vspace{-0.35cm}
\paragraph{Context-aware\&\&Shared weights.}As shown in Tab.~\ref{tab:abattn}, we investigated the model's performance when combining the global branch with either shared weights or context-aware weights. When using only shared weights to extract high-frequency local information, the Top1 accuracy of classification is 1.8\% lower compared to using the complete AttnConv. Similarly, when using only context-aware weights, the result is 0.6\% lower. These results fully demonstrate the rationality of AttnConv.
\vspace{-0.35cm}
\paragraph{Context-aware Weights Generation.}We compared our method for context-aware weights generation with local self-attention (window self-attention), conducting experiments on three approaches of local feature extraction, as shown in Fig.~\ref{fig:comp}. These approaches include: (a) Standard convolution that only uses shared weights; (b) a combination of context-aware weights and shared weights, where context-aware weights are calculated by window self-attention; and (c) our AttnConv. As shown in Tab.~\ref{tab:3comp}, AttnConv achieves performance improvement at the cost of a slight increase in the number of parameters. The gradual transformation from convolution to AttnConv verifies the validity of AttnConv. Specifically, compared to (b) in Fig.~\ref{fig:comp}, our AttnConv achieves an advantage of 1.2\% on top1 accuracy, \textbf{+1.1} on box AP, \textbf{+1.2} on mask AP, and \textbf{+0.9} on mIoU. More experiments can be found in \textcolor{red}{Appendix}.
\vspace{-0.05cm}
\section{Conclusion}
\vspace{-0.05cm}
In this paper, we propose CloFormer, a lightweight vision transformer with context-aware local enhancement, and develop a novel method for local perception. Our CloFormer achieves competitive performance among models with similar FLOPs and model sizes. Moreover, the carefully designed AttnConv extracts high-frequency local representations effectively by making full use of the shared weights and context-aware weights. In addition, we use a two-branch structure to mix the high- and low-frequency information. The AttnConv and the structure enable our model to excel in various visual tasks. Extensive experiments demonstrate that CloFormer is a solid, lightweight vision backbone that outperforms many existing backbones.

{\small
\bibliographystyle{ieee_fullname}
\bibliography{egbib}
}
\clearpage
\begin{appendices}
\section{Implementation Details}
\subsection{Architecture Details}

The detailed architectures are shown in Tab.~\ref{tab:acr}, where all variants use the input size of $224^2$. In the three variants of CloFormer, we only modify the channel dimensions of the stages, while the depth of the four stages in different variants are fixed at [2, 2, 6, 2]. We measure the FLOPs for image classification at resolution and $224\times224$, where we use a global average pooling layer and a fully-connected layer to get the output. 

\subsection{Image Classification}

We follow the same training strategy with DeiT~\cite{deit}. All models are trained for 300 epochs from scratch. The AdamW optimizer with a cosine decay learning rate scheduler is used to train the model. Moreover, the initial learning rate, weight decay, and batch size are 0.001, 0.05, and 1024, respectively. Our augmentation settings are RandAugment~\cite{randomaugment} (randm9-mstd0.5-inc1), Mixup~\cite{mixup} (prob=0.8), CutMix~\cite{cutmix} (prob=1.0), Random Erasing~\cite{randera} (prob=0.25), increasing stochastic depth~\cite{droppath} (prob=0, 0.06, 0.06 for CloFormer-XXS, CloFormer-XS, and CloFormer-S, respectively).

\subsection{COCO Object Detection}

Based on MMDetection~\cite{mmdetection}, we use Mask-RCNN~\cite{maskrcnn} and RetinaNet~\cite{retinanet} to conduct object detection and instance segmentation. All models are trained with "$1\times$" schedules (12 epochs), and images are resized to the shorter side of 800 pixels while the longer side is within 1333 pixels. The AdamW optimizer is used with a learning rate of 0.0001, weight decay of 0.0001, and the total batch size is set to 16. Similar to other works~\cite{ortho, litv1, litv2}, the learning rate in our models declines with the decay rate of 0.1 at epoch 8 and epoch 11. 

\subsection{ADE20K Semantic Segmentation}

The well-known semantic segmentation framework: Semantic FPN~\cite{semanticfpn} is adopted to evaluate our backbones. We follow the same setting with PVT~\cite{pvt} and EdgeViT~\cite{edgevit}, and use the "$1\times$" schedule (trained for 80k iterations). All the models are trained with the input resolution of $512\times512$. When testing the model, we resize the shorter side of the image to 512 pixels.

\section{More Ablation and Analysis Results}

\subsection{Ablation of $K$ in AttnConv}

In AttnConv, we computed the Hadamard product of $Q$ and $K$ during the process of generating context-aware weights. Here, we eliminate $K$ and only use $Q$ to generate context-aware weights. The results are shown in Tab.~\ref{tab:ab_k}. We found from the experimental results that using $Q\odot K$ achieved a performance advantage of \textbf{+0.6} compared to using $Q$ alone while incurring only a negligible increase in the number of parameters and FLOPs.
\begin{table}[h]
    \centering
    \begin{tabular}{c|c|c|c}
    \hline
         Model & Params(M) & FLOPs(G) & Top-1 acc(\%)\\
         \hline
         only $Q$& 4.0 & 0.60 & 76.4\\
         $Q\odot K$ & 4.2 & 0.62 & \textbf{77.0} \\
         \hline
    \end{tabular}
    \caption{Ablation of $K$.}
    \label{tab:ab_k}
\end{table}

\subsection{Nonlinearity in Context-aware Generation}

\begin{figure}[h]
    \centering
    \includegraphics[scale=0.33]{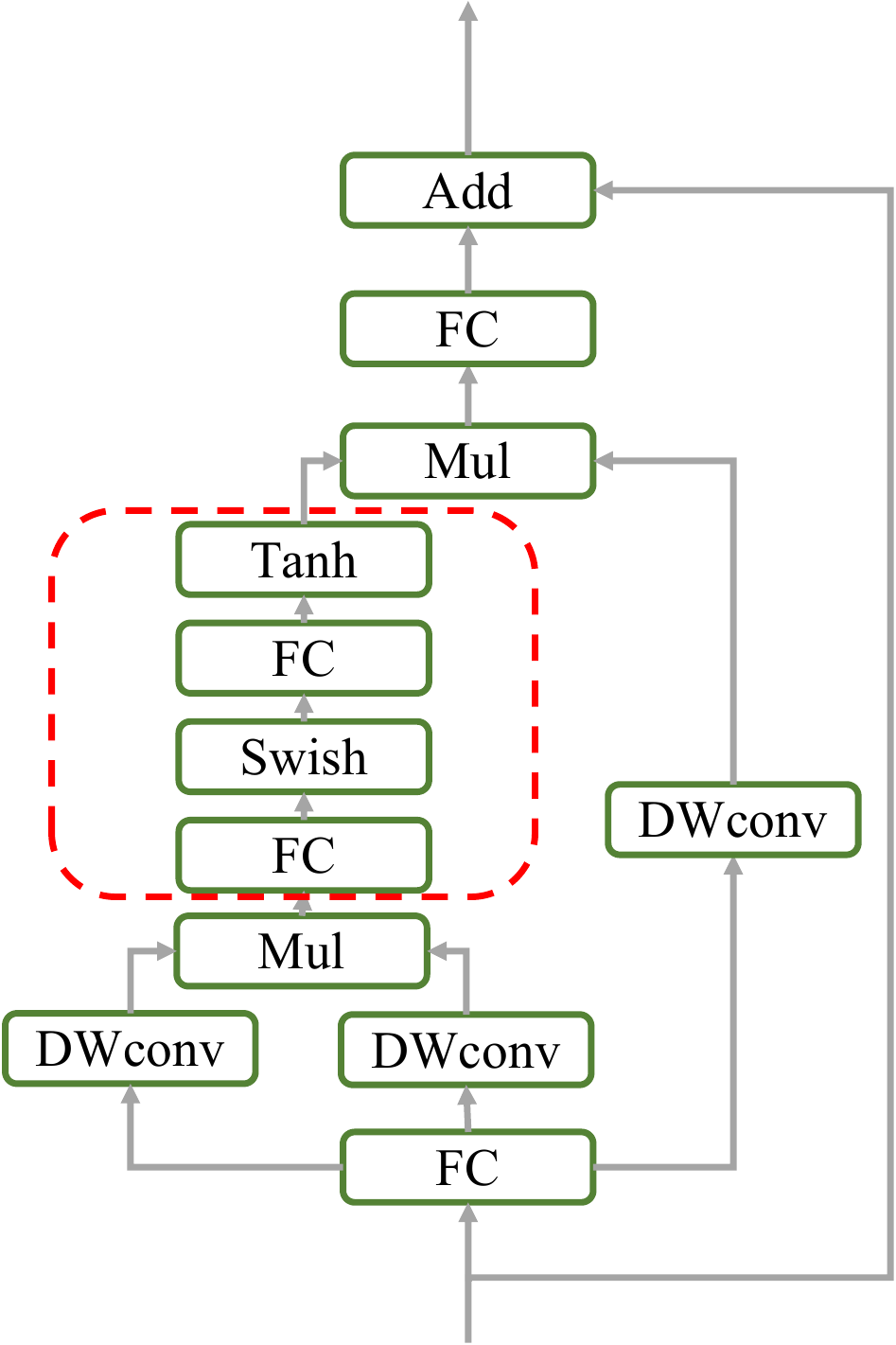}
    \caption{Illustration of our nonlinearity component.}
    \label{fig:ab_non}
\end{figure}

\begin{figure*}[h]
    \centering
    \includegraphics[scale=0.34]{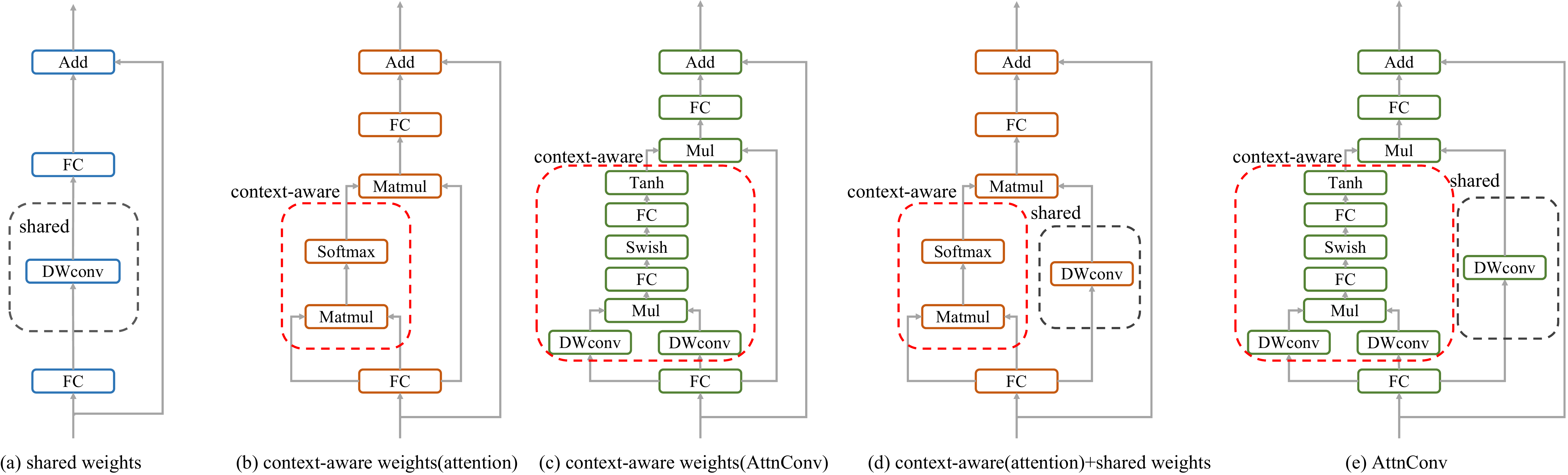}
    \caption{Different methods for local perception. }
    \label{fig:local_percp}
\end{figure*}

In AttnConv, we have introduced stronger non-linearity compared to commonly used attention mechanisms during the process of generating context-aware weights. In this section, we attempt to introduce even stronger non-linearity than that in the main text and have found that the model's performance still improves. As shown in Fig.~\ref{fig:ab_non}, our nonlinear component consists of $\bm{{\rm Swish}}$ and $\bm{{\rm Tanh}}$, connected by a fully-connected layer. In this part, we analyzed the performance of the model under different scenarios, including using only $\bm{{\rm Tanh}}$, adding one $\bm{{\rm Swish}}$, adding two $\bm{{\rm Swish}}$, and adding three $\bm{{\rm Swish}}$. The experimental results are shown in Tab.~\ref{tab:ab_non}. We found that after incorporating additional $\bm{{\rm Swish}}$, there was still a performance improvement in the classification results.

\begin{table}[h]
    \centering
    \begin{tabular}{c|c|c|c}
    \hline
        Model & Params(M) & FLOPs(G) & Top-1 acc(\%)\\
        \hline
         only $\bm{{\rm Tanh}}$& 4.1 & 0.60 & 76.4 \\
         \hline
         +$\bm{{\rm FC}}$& 4.2 & 0.62 & 76.5 \\
         +$\bm{{\rm Swish}}$& 4.2 & 0.62 & 77.0 \\
         \hline
         +$\bm{{\rm FC}}$& 4.2 & 0.63 & 77.0 \\
         +$\bm{{\rm Swish}}$& 4.2 & 0.63 & 77.2 \\
         \hline
         +$\bm{{\rm FC}}$& 4.2 & 0.64 & 77.2 \\
         +$\bm{{\rm Swish}}$& 4.2 & 0.64 & \textbf{77.3} \\
         \hline
    \end{tabular}
    \caption{Ablation of nonlinearity components.}
    \label{tab:ab_non}
\end{table}

\subsection{Different Methods for Local Perception}

To demonstrate the superiority of our proposed local perception module——AttnConv; in this section, we extensively investigate the impact of various local perception methods on the model's performance. As shown in Fig.~\ref{fig:local_percp}, we compare five methods for local perception: (a) only use shared weights. (b) only use context-aware weights generated by local self-attention (window self-attention). (c) only use context-aware weights generated by our method. (d) Use shared and context-aware weights, where the context-aware weights are generated by local self-attention (window self-attention). (d) our AttnConv. All five configurations utilized a global branch and the same channel split setting. The results are shown in Tab.~\ref{tab:ab_dif}.

\begin{table}[h]
    \centering
    \begin{tabular}{c|c|c|c}
    \hline
         Model & Params(M) & FLOPs(G) & Top-1 acc(\%)\\
         \hline
         (a) & 3.9 & 0.55 & 75.2 \\
         (b) & 4.0 & 0.60 & 75.3\\
         (c) & 4.1 & 0.62 & 76.4 \\
         (d) & 4.0 & 0.61 & 75.8 \\
         (e) & 4.2 & 0.62 & \textbf{77.0}\\
         \hline
         
    \end{tabular}
    \caption{Comparison among five different methods.}
    \label{tab:ab_dif}
\end{table}

\subsection{Ablation of Activation Function}

In this section, we explored the impact of different activation functions in Attnconv on model performance, as shown in Tab.~\ref{tab:ab_act}. It can be seen that the combination of $\bm{{\rm Swish}}$ and $\bm{{\rm Tanh}}$ gives our model the highest performance.
\begin{table}[h]
\vspace{-0.35cm}
    \centering
    \setlength{\tabcolsep}{0.4mm}
    \begin{tabular}{c|c|c|c}
         \hline
         Model & Params(M) & FLOPs(G) & Top-1 acc(\%)\\
         \hline
         $\bm{{\rm GELU}}+\bm{{\rm Tanh}}$ & 4.2 & 0.62 & 76.8\\
         $\bm{{\rm SiLU}}+\bm{{\rm Tanh}}$ & 4.2 & 0.62 & 76.9\\
         $\bm{{\rm ReLU}}+\bm{{\rm Tanh}}$ & 4.2 & 0.62 & 76.7\\
         \hline
         $\bm{{\rm Swish}}+\bm{{\rm GELU}}$ & 4.2 & 0.62 & 76.7\\
         $\bm{{\rm Swish}}+\bm{{\rm Sigmoid}}$ & 4.2 & 0.62 & 76.8\\
         %$\bm{{\rm Swish}}+None$ & 4.2 & 0.62 & \\
         \hline
         $\bm{{\rm Swish}}+\bm{{\rm Tanh}}$ & 4.2 & 0.62 & \textbf{77.0}\\
         \hline
    \end{tabular}
    \caption{Ablation of activation function in AttnConv.}
    \label{tab:ab_act}
\end{table}

\subsection{Ablation of Early Conv}

We compare the performance of using conv stem versus direct patch embedding, as illustrated in Tab.~\ref{tab:ab_early}. Utilizing conv stem yields a performance gain of \textbf{+0.5} for the model.

\begin{table}[h]
    \centering
    \begin{tabular}{c|c|c|c}
    \hline
         Model & Params(M) & FLOPs(G) & Top-1 acc(\%) \\
         \hline
         Patch embed & 4.1 & 0.55 & 76.5\\
         Conv stem & 4.2 & 0.62 & \textbf{77.0}\\
         \hline
    \end{tabular}
    \caption{Ablation of early conv.}
    \label{tab:ab_early}
\end{table}

\begin{table*}[!htbp]
    \centering
    \setlength{\tabcolsep}{4mm}
    \begin{tabular}{c|c|c|c|c|c}
    \toprule[1pt]
        \multirow{2}{*}{Stage} & \multirow{2}{*}{Output Size} & \multirow{2}{*}{Layer Settings} & \multicolumn{3}{c}{CloFormer}\\
        \cline{4-6}
        & & & XXS & XS & S\\
        \midrule[0.5pt]
        \multirow{5}{*}{Patch embed} & \multirow{5}{*}{$\frac{H}{4}\times\frac{W}{4}$} & \multirow{5}{*}{Conv stem} & \multicolumn{3}{c}{Conv{$3\times3$}, stride 2, C/2}\\
         & & & \multicolumn{3}{c}{Conv{$3\times3$}, stride 2, C}\\
         & & & \multicolumn{3}{c}{Conv{$3\times3$}, stride 1, C}\\
         & & & \multicolumn{3}{c}{Conv{$3\times3$}, stride 1, C}\\
         & & & \multicolumn{3}{c}{Conv{$1\times1$}, stride 1, C}\\
         \midrule[0.5pt]
        \multirow{6}{*}{Stage1} & \multirow{6}{*}{$\frac{H}{4}\times\frac{W}{4}$} & Embed dim & 32 & 48 & 64 \\
        \cline{3-6}
        & & Channel split ([local, global]) & [24, 8] & [32, 16] & [48, 16]\\
        \cline{3-6}
         & & AttnConv kernel size & 3 & 3 & 3\\
         \cline{3-6}
         & & Pooling stride & 8 & 8 & 8 \\
         \cline{3-6}
         & & ConvFFN kernel size & 5 & 5 & 5 \\
         \cline{3-6}
         & & ConvFFN ratio & 4 & 4 &4 \\
         \midrule[0.5pt]
         \multirow{6}{*}{Stage2} & \multirow{6}{*}{$\frac{H}{8}\times\frac{W}{8}$} & Embed dim & 64 & 96 & 128 \\
        \cline{3-6}
        & & Channel split ([local, global]) & [32, 32] & [48, 48] & [64, 64]\\
        \cline{3-6}
         & & AttnConv kernel size & 5 & 5 & 5\\
         \cline{3-6}
         & & Pooling stride & 4 & 4 & 4 \\
         \cline{3-6}
         & & ConvFFN kernel size & 5 & 5 & 5 \\
         \cline{3-6}
         & & ConvFFN ratio & 4 & 4 &4 \\
         \midrule[0.5pt]
         \multirow{6}{*}{Stage3} & \multirow{6}{*}{$\frac{H}{16}\times\frac{W}{16}$} & Embed dim & 128 & 160 & 224 \\
        \cline{3-6}
        & & Channel split ([local, global]) & [64, 64] & [80, 80] & [112, 112]\\
        \cline{3-6}
         & & AttnConv kernel size & 7 & 7 & 7\\
         \cline{3-6}
         & & Pooling stride & 2 & 2 & 2 \\
         \cline{3-6}
         & & ConvFFN kernel size & 5 & 5 & 5 \\
         \cline{3-6}
         & & ConvFFN ratio & 4 & 4 &4 \\
         \midrule[0.5pt]
         \multirow{6}{*}{Stage4} & \multirow{6}{*}{$\frac{H}{32}\times\frac{W}{32}$} & Embed dim & 256 & 352 & 448 \\
        \cline{3-6}
        & & Channel split ([local, global]) & [64, 192] & [112, 240] & [112, 336]\\
        \cline{3-6}
         & & AttnConv kernel size & 9 & 9 & 9\\
         \cline{3-6}
         & & Pooling stride & 1 & 1 & 1 \\
         \cline{3-6}
         & & ConvFFN kernel size & 5 & 5 & 5 \\
         \cline{3-6}
         & & ConvFFN ratio & 4 & 4 &4 \\
         \midrule[0.5pt]
         Classifier & 1000 & \multicolumn{4}{c}{Global Average Pooling \&\& Linear}\\
         \midrule[0.5pt]
         \multicolumn{3}{c|}{Params(M)} & 4.2 & 7.2 & 12.3 \\
         \midrule[0.5pt]
         \multicolumn{3}{c|}{FLOPs(G)} & 0.6 & 1.1 & 2.0 \\
         \bottomrule[1pt]
    \end{tabular}
    \caption{Details about CloFormer's architecture.}
    \label{tab:acr}
\end{table*}

\end{appendices}

\end{document}